\documentclass{article}

\usepackage[preprint]{neurips_2019}

\usepackage[utf8]{inputenc} 
\usepackage[T1]{fontenc}    
\usepackage{hyperref}       
\usepackage{url}            
\usepackage{booktabs}       
\usepackage{amsfonts}       
\usepackage{nicefrac}       
\usepackage{microtype}      
\usepackage{graphicx,wrapfig}
\usepackage[skip=2pt,font=footnotesize,labelfont=bf]{caption}
\usepackage{subcaption}
\usepackage{amsmath,amssymb}
\usepackage{array}
\usepackage{wrapfig}

\newcolumntype{U}{>{\centering\arraybackslash}p{2.90cm}}
\newcolumntype{V}{>{\centering\arraybackslash}p{2.00cm}}
\newcolumntype{T}{>{\centering\arraybackslash}p{1.0cm}}

\title{Reachability Embeddings: Scalable Self-Supervised Representation Learning from Spatiotemporal Motion Trajectories for Multimodal Computer Vision}

\author{%
    Swetava Ganguli\thanks{Corresponding author. Alternative EMail: \texttt{swetava@cs.stanford.edu}}, C. V. Krishnakumar Iyer, Vipul Pandey \\
    Apple\\
    \texttt{\{swetava,cvk,vipul\}@apple.com} \\
}

\begin{document}

\maketitle

\setlength{\belowcaptionskip}{0pt}
\setlength{\textfloatsep}{0pt}
\begin{wrapfigure}{L}{0.50\textwidth}
    \centering
    \includegraphics[width=0.50\textwidth]{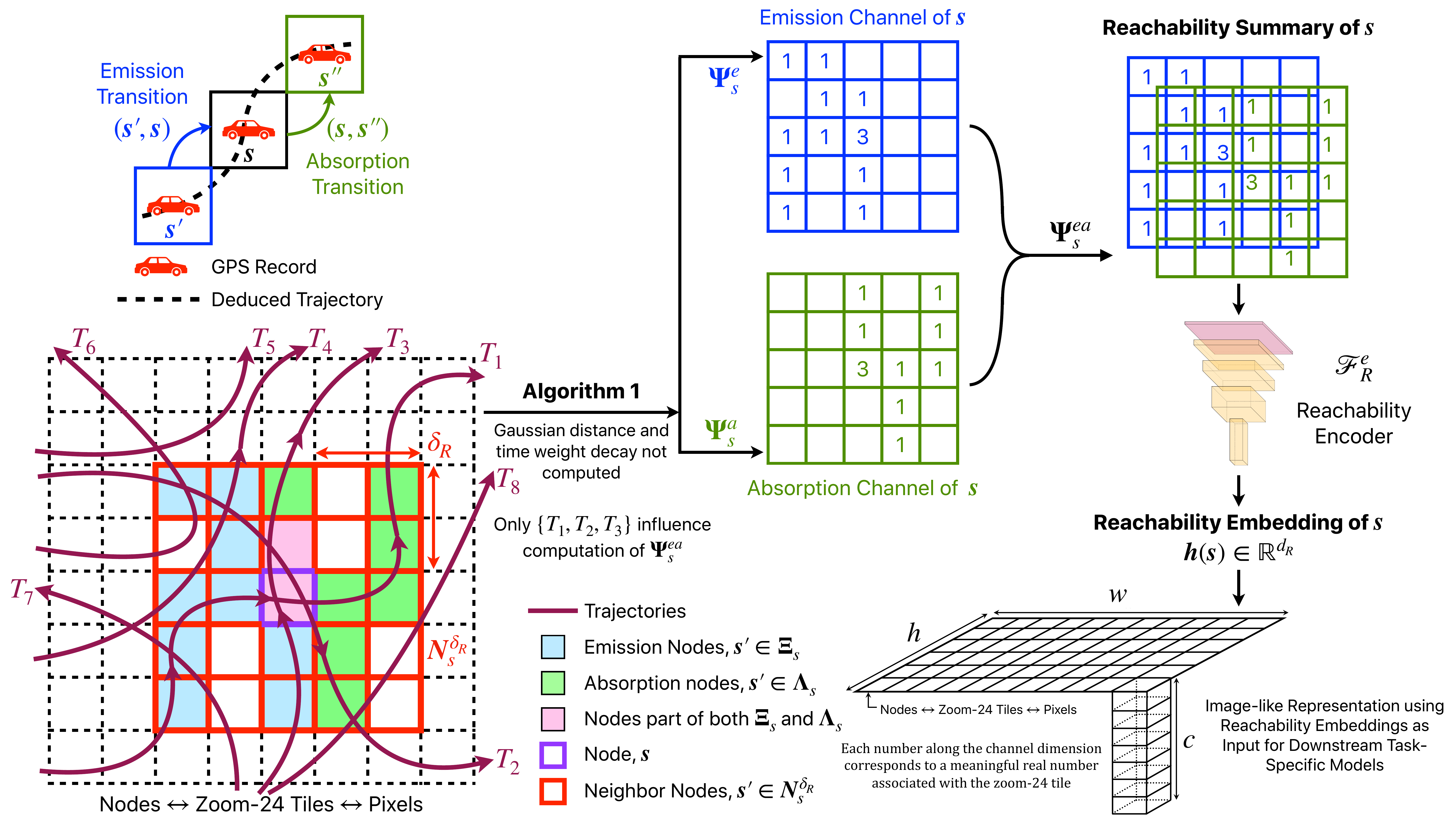}
    \caption{Generation of reachability embeddings.}
    \label{reachability_generation}
\end{wrapfigure}

\vspace{-1.0cm}
\section{Introduction}\label{section::introduction}
\vspace{-0.3cm}
Graphs are natural data structures for representing geospatial datasets (e.g., road networks, point clouds, 3D object meshes) with natural definitions of nodes and edges. Instead of hand-engineering task-specific and domain-specific features for nodes in graphs, recent methods \cite{metapath2vec,node2vec,deepwalk} have focused on automatically learning low-dimensional, feature vector representations called node embeddings. In parallel, self-supervised learning has been an area of active and promising research. Self-supervised representation learning techniques utilize large datasets without semantic annotations to learn meaningful, universal features that can be conveniently transferred to solve a wide variety of downstream supervised tasks.
In \cite{ganguli2021reachability}, we propose a novel self-supervised method for learning representations of geographic locations from observed unlabeled GPS trajectories that can be used by itself or can be combined with other image-like data modalities to solve downstream geospatial computer vision tasks. A spatial proximity-preserving graph representation of the earth surface is inferred from observed mobility trajectories that is used to cast the geospatial representation learning task into a task of learning self-supervised node embeddings. Reachability embeddings serve as task-agnostic, feature representations of geographic locations. Using reachability embeddings as pixel representations for five different downstream geospatial tasks, cast as supervised semantic segmentation problems, we quantitatively demonstrate that reachability embeddings are semantically meaningful representations and result in 4--23\% gain in performance, as measured using area under the precision-recall curve (AUPRC) metric, when compared to baseline models that use pixel representations (called \textit{Local Aggregate Representations} (LAR)) that do not account for the spatial connectivity between tiles \cite{syntheticdata}. Reachability embeddings transform sequential, spatiotemporal motion trajectory data into semantically meaningful image-like tensor representations that can be combined (multimodal fusion) with other data modalities (on machine learning platforms such as Trinity \cite{trinity}) that are or can be transformed into image-like tensor representations (for e.g., RBG imagery, graph embeddings of road networks, passively collected imagery like SAR, etc.) to facilitate multimodal learning in geospatial computer vision. Multimodal computer vision is critical for training machine learning models for geospatial feature detection to keep a geospatial mapping service up-to-date in real-time and can significantly improve user experience and above all, user safety. 


\section{The Reachability Embeddings Algorithm Proposed in \cite{ganguli2021reachability}}\label{section::reachabilityembeddings}
\vspace{-0.2cm}
A \textit{GPS trajectory} encodes spatiotemporal movement of an object as a chronologically ordered sequence of GPS records (a tuple of timestamp and the location's zoom-24 tile \cite{z24tiledefinition}). Let $\mathcal{T}$ represent the set of all available GPS trajectories during the time interval $[t_0, t_0+\Delta t]$ such that all GPS records in each trajectory are associated with the same motion modality (e.g., driving, walking, biking). The \textit{Earth Surface Graph} (ESG), $\boldsymbol{G}_{ES}(\boldsymbol{V}_{ES},\boldsymbol{E}_{ES})$, is defined as the inferred graph obtained from a raster representation of the earth's surface based on zoom-24 tiles with these tiles as nodes, $\boldsymbol{V}_{ES}$. 
\setlength{\belowcaptionskip}{0pt}
\setlength{\textfloatsep}{0pt}
\begin{wrapfigure}{R}{0.19\textwidth}
    \centering
    \includegraphics[width=0.19\textwidth]{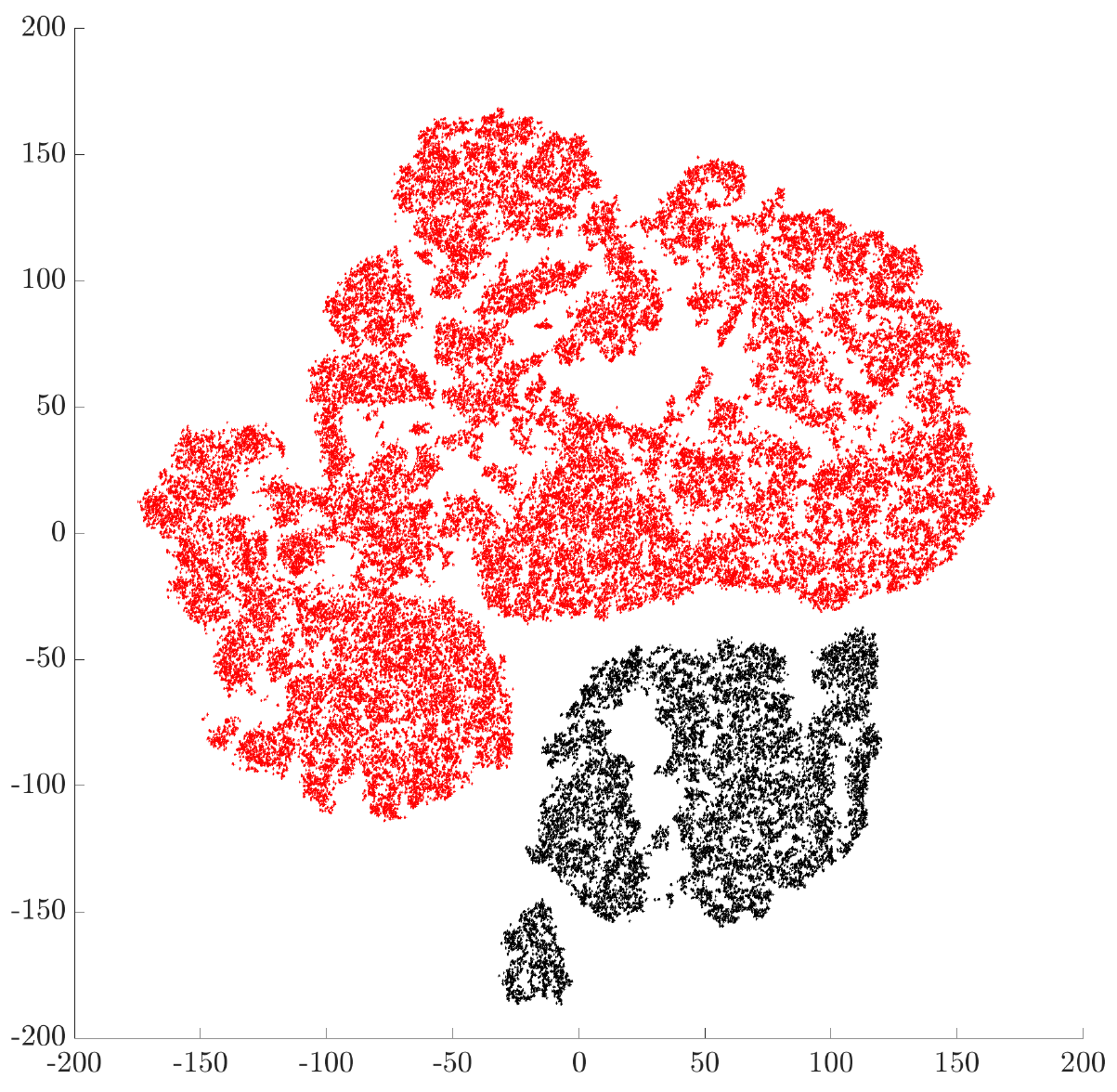}
    \caption{UMAP of embeddings: Highways (black), otherwise (red).}
    \label{umap}
\end{wrapfigure}
GPS trajectories are modeled as allowed Markovian paths on the ESG thereby defining the edges of the ESG, $\boldsymbol{E}_{ES}$, as the observed transitions between nodes in $\mathcal{T}$. In summary, the following equivalences are defined: (i) zoom-24 tile $\leftrightarrow$ Markovian state space $\leftrightarrow$ $\boldsymbol{V}_{ES}$, (ii) GPS trajectory $\leftrightarrow$ Markovian trajectory in $\boldsymbol{V}_{ES}$ $\leftrightarrow$ Path in $\boldsymbol{G}_{ES}$. In \cite{ganguli2021reachability}, a two stage algorithm is proposed to learn self-supervised representations for each node, $\boldsymbol{s}$, in the ESG based on the frequency of transitions from neighboring nodes to $\boldsymbol{s}$ (\textit{emission transitions}) and transitions from $\boldsymbol{s}$ to its neighboring nodes (\textit{absorption transitions}). With the intuition that transitions occurring over large spatial distances should not influence the learned representations of nodes, transitions occurring only within a user-specified neighborhood ($\boldsymbol{N}_{s}$) are considered valid. The first stage comprises of a scalable, distributed, and data-parallel algorithm that exploits the Cartesian grid structure of $\boldsymbol{V}_{ES}$ to generate spatial proximity-preserving, image-like, tensor representations, called the \textit{reachability summary}, of the transitions associated with a node (Figure \ref{reachability_generation}) in $\mathcal{T}$ for every node $\boldsymbol{s} \in \boldsymbol{V}_{ES}$. In the second stage, reachability summaries are compressed to a $\boldsymbol{d}_{R}$-dimensional vector representation for each tile/node, called \textit{reachability embeddings}, using the encoder of a contractive, fully-convolutional autoencoder trained to reconstruct reachability summaries. Reachability summary generation and its contractive reconstruction can together be viewed as the self-supervision pretext task to obtain reachability embeddings. The contractive regularization incentivizes robustness and invariance of embeddings to small perturbations in the summary leading embeddings of geographic locations with similar mobility activity to be close to each other (Figure \ref{umap}) in the learned, low-dimensional manifold. Reachability summary generation is the expensive stage influencing the runtime of the algorithm at test time. We demonstrate the scalability of the proposed distributed, data-parallel algorithm for stage 1 using strong scaling analysis \cite{grama03} on the publicly available T-Drive dataset \cite{msrtdrive}. Reachability summaries are shown to capture spatial connectivity patterns based on frequency, distance traveled, and time taken during transitions. A theoretical motivation of reachability summaries and reachability embeddings using the Chapman-Kolmogorov Equations (CKE) for Markov chains is also provided in \cite{ganguli2021reachability}. Reachability embeddings are designed to be easy for storage and management by ML feature stores for efficient multimodal modeling of downstream tasks. We use the Trinity Feature Store and the Trinity ML platform \cite{trinity} to store the generated embeddings and for the experiments described in the next section.

\section{Qualitative and Quantitative Results}\label{section::results}
\vspace{-0.3cm}
The impact of reachability embeddings on the performance of downstream tasks is evaluated by training supervised, pixel-wise, semantic segmentation models for five important downstream geospatial tasks, viz. (i) detection of overpasses, (ii) detection of pedestrian crosswalks, (iii) detection of driving access (entry/exit) points, (iv) detection of locations with traffic lights, and (v) detection of locations with stop signs. For all tasks, the UNet \cite{ronneberger2015} architecture is trained to minimize the pixel-wise binary cross-entropy between the predicted segmentation map and labels. A 60-20-20 randomly chosen training-validation-testing set is created and fixed for all subsequent experimentation. Table \ref{results_table} shows the quantitative comparison of the performance for all 5 downstream tasks, quantified using AUPRC on the test set after model training converges (100 epochs), between two variants each of baseline (varying observation intervals) and reachability-based (varying embedding dimension) models. Precision-Recall curves for all tasks will be shown. Three key observations emerge: (i) increasing the observation interval (increasing signal-to-noise ratio) for computing LAR and increasing $\boldsymbol{d}_R$ increases AUPRC --- for most values of recall, precision increases due to reduction in false positives; (ii) reachability-based models outperform the baseline models, including those using LAR computed by observing 3 times more trajectories: the AUPRC gain by the inferior reachability-based model ($\boldsymbol{d}_R = 8$) over the superior baseline model (observation interval $3\Delta t$) varies from 1.6\% for stop signs detection task to 18.4\% for the access point detection task; (iii) for the same observation time interval, $\Delta t$, simply replacing the LAR-based inputs by reachability embeddings ($\boldsymbol{d}_R = 16$) results in an AUPRC gain that varies from 4.1\% for the stop signs detection task to 23.3\% for the access point detection task. These observations conclusively demonstrate that reachability embeddings are more informative, denser representations of trajectory data requiring lesser trajectories (upto 67\% less) to compute. Thus, reachability embeddings  may be used to compute semantically meaningful representations of trajectories in geographical areas with less traffic or to build computer vision-based models for low-resource geospatial tasks. Additional results of using reachability embeddings in multimodal settings (combining with satellite imagery and road network graph via early fusion) and qualitative comparison of model predictions using reachability embeddings as inputs \cite{ganguli2021reachability} will be shown.

\begin{table}[ht]
    \footnotesize
    \caption{Comparison of AUPRC ($\uparrow$ is better) obtained from both variants of LAR and reachability-based models for 5 downstream geospatial tasks. Percentage gain compared to first row shown in brackets. \label{results_table}}
    \centering
    \begin{tabular}{U T V V V V V}
        \toprule
        Input Channels                      & Observation Interval & Overpass Detection       & Crosswalk Detection     & Access Point Detection   & Traffic Lights Detection & Stop Signs Detection     \\
        \midrule
        LAR Baseline                        & $\Delta t$           & 0.782                    & 0.922                   & 0.663                    & 0.890                    &  0.921                   \\  
        LAR Baseline                        & $3\Delta t$          & 0.785 (+0.4\%)           & 0.923 (+0.1\%)          & 0.684 (+3.0\%)           & 0.925 (+3.8\%)           &  0.924 (+0.3\%)          \\ 
        Reachability, $\boldsymbol{d}_R=8$  & $\Delta t$           & 0.899 (+15.0\%)          & 0.959 (+4.0\%)          & 0.843 (+21.4\%)          & 0.974 (+8.6\%)           &  0.938 (+1.9\%)          \\ 
        Reachability, $\boldsymbol{d}_R=16$ & $\Delta t$           & \textbf{0.925} (+18.3\%) & \textbf{0.976} (+5.9\%) & \textbf{0.864} (+23.3\%) & \textbf{0.985} (+9.6\%)  &  \textbf{0.959} (+4.1\%) \\
        \bottomrule
    \end{tabular}
\end{table}

\clearpage
\bibliographystyle{abbrv}
\bibliography{Paper_BayLearn2022}

\end{document}